\pdfoutput=1

\documentclass[11pt]{article}

\usepackage{booktabs}
\usepackage{multirow}

\usepackage[]{acl}

\usepackage{times}
\usepackage{latexsym}

\usepackage[T1]{fontenc}

\usepackage[utf8]{inputenc}

\usepackage{microtype}

\usepackage{graphicx}

\usepackage{caption}
\usepackage{subcaption}

\usepackage{tabularx}
\usepackage{enumitem}

\newcommand{\vl}{V\&L }

\title{Understanding  Cross-modal Interactions in \vl Models that Generate Scene Descriptions}

\author{Michele Cafagna$^1$ \ \ ~~Kees van Deemter$^2$ \ \ ~~Albert Gatt$^{1,2}$\\\\
$^1$University of Malta, Institute of Linguistics and Language Technology \\
$^2$Universiteit Utrecht, Information and Computing Sciences \\ 
\texttt{michele.cafagna@um.edu.mt}\\\texttt{\{a.gatt, c.j.vandeemter\}@uu.nl}
}

\begin{document}
\maketitle
\begin{abstract}
Image captioning models tend to describe images in an object-centric way, emphasising visible objects. But image descriptions can also abstract away from objects and describe the type of scene depicted. In this paper, we explore the potential of a state of the art Vision and Language model, VinVL, to caption images at the scene level using  (1) a novel dataset which pairs images with both object-centric and scene descriptions. Through (2) an in-depth analysis of the effect of the fine-tuning, we show (3) that 
a small amount of curated
data suffices to generate scene descriptions without losing the capability to identify object-level concepts in the scene;  
the model acquires a more holistic view of the image compared to when object-centric descriptions are generated.
We discuss the parallels between these results and insights from computational and cognitive science research on scene perception.
\end{abstract}

\section{Introduction}
\label{sec:intro}

\begin{figure}[!t]
    \centering
    \includegraphics[scale=0.25]{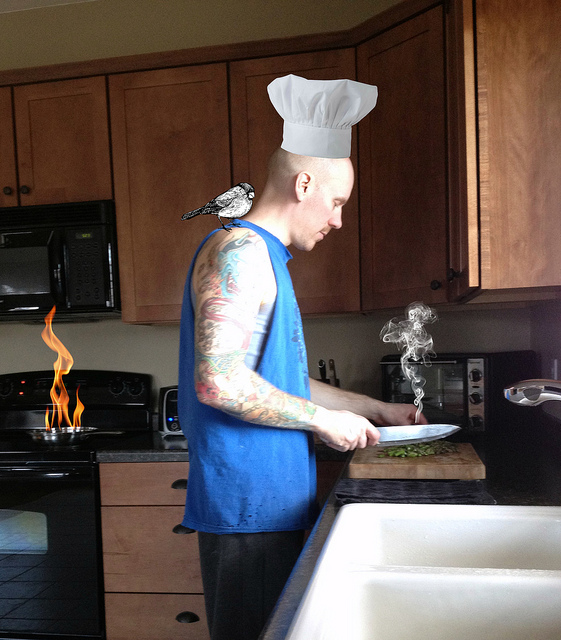}
    \caption{\small Image from the MS-COCO 2014 validation set. One reference caption is: {\em a man in a chefs hat chopping food}.}
    \label{fig:coco-ex}
\end{figure}

When humans view images, they can quickly capture their `gist'. For example, it is immediately evident that Figure~\ref{fig:coco-ex} is a kitchen. Such judgments are fast and are informed by expectations about which objects occur in typical scenes (`scene semantics') and their configuration (`syntax') \cite{Malcolm2016,Vo2021,Self2019}. This knowledge affects the deployment of attentional resources \cite{Torralba2006,Oliva2007,Wu2014,Henderson2017}.
Scene understanding and object recognition constrain the selection of attended locations in human visual attention \cite{itti2001computational}.

In this paper, we explore the implications of these findings for image captioning models. 
There are at least two levels at which an image can be appraised. An {\bf object-centric} perspective focuses primarily on individual objects and actions (e.g. the example caption in Fig~\ref{fig:coco-ex}). This has dominated captioning models \cite[see][for an early, influential statement of this view]{Hodosh2013} and has informed the design of widely-used datasets, which pair images with captions that explicitly mention at least some of the objects in a picture \cite[e.g.][]{young-etal-2014-image,Chen2015,pont2020connecting,Gurari2019VizWizPrivAD,Sharma2018,agrawal2019nocaps}. In contrast, a {\bf scene-level} caption (e.g. `a kitchen' for Figure~\ref{fig:coco-ex}) contains
less object-specific detail. Such captions are less redundant with respect to the image they describe, but convey enough information to generate inferences about content and structure (e.g. kitchens typically contain cupboards, but not birds; etc). 

Most image captioning datasets contain object-centric captions and no currently available resource pairs both scene-level and object-centric captions with images. In this paper, we address this gap and ask (i) whether 
captioning models can be adapted both for object-centric and scene-level captioning and (ii) whether 
the two strategies rely on different types of interplay between the visual and linguistic modalities. 
Addressing these questions
can shed light on the ability of \vl models to reason about the relationship between scenes and their components. In addition, it is desirable for models to generate scene-level descriptions as well as object-centric ones. In many communicative contexts, scene-level captions are informative and non-redundant,
recalling the quality and the quantity discourse maxims defined by \citet{Grice1975}.

We present a study of object-centric versus scene-level captioning. We focus on VinVL \cite{zhang2021vinvl}, a BERT-based model in the OSCAR family \cite{li2020oscar} of models, which have recently dominated the state of the art in image captioning.\footnote{ At the time of this work, three OSCAR-based models (OSCAR, VinVL, {\sc lemon}) are among the top 5 in the \href{https://paperswithcode.com/sota/image-captioning-on-coco-captions?metric=CIDER}{leaderboard} of the COCO image captioning task.}
Our main contributions are:
\begin{enumerate}[label=\roman{*})]
    \item We introduce a novel dataset, HL-Scenes (Sec~\ref{sec:data}) extending part of the COCO dataset \cite{Chen2015} with scene-level descriptions.
    \item 
    We perform an in-depth investigation of the impact of fine-tuning on the pre-trained model. The analysis is designed to thoroughly inspect object-scene relations by exploiting cross-modal attention (Sec~\ref{sec:attention}), coupled with probing (Sec~\ref{sec:prob}) and ablation studies (Sec~\ref{sec:ablation}). 
    \item  We show that (i) VinVL's pre-trained representations are rich enough to support scene-level captioning, but that (ii) fine-tuning results in a different deployment of attentional resources. This bears parallels to the findings in research on human scene perception.
\end{enumerate}

\section{Related work}
\label{sec:rel_work}

\paragraph{Datasets}
Existing image-caption datasets emphasise object-centric captions \cite[an early exception, using abstract scenes, is][]{Ortiz2015}. This is also true of web-sourced datasets such as Conceptual Captions \cite[CC;][]{Sharma2018}.
For example, the CC filtering pipeline explicitly checks for overlaps between caption tokens and objects identified in the image. The \texttt{nocaps} benchmark \cite{agrawal2019nocaps} tests models' ability to generalise to 
out-of-domain objects. There are several \vl datasets and tasks which introduce knowledge-rich annotations and address models' ability to reason with linguistic and visual cues \cite{zellers2019recognition,Zellers2018,Suhr2017,Suhr2019,Park2020,Pezzelle2020}. In this paper, we take this line of work further by introducing the novel HL-Scenes dataset, which pairs object-centric and scene-level captions to images.

\paragraph{Models}
 Transformer-based \vl models are usually divided into 
\textit{single-stream}
\cite{li2019visualbert,chen2019uniter, li2020oscar,su2019vl}
and \textit{dual-stream} 
\cite{tan2019lxmert, lu2019vilbert,radford2021learning} architectures. It has been shown that single- and dual- stream models perform roughly at par under the same training settings \cite{bugliarello2021multimodal}. On the other hand \citet{hendricks2021decoupling} showed that model performance is highly impacted by dataset curation, attention, and loss function definition.



Most \vl single-stream models are inspired by BERT \cite{devlin2018bert}. They incorporate the visual modality in the form of features extracted using 
a visual backbone, 
typically a
Faster-RCNN \cite{ren2015faster} pre-trained on an object labelling task such as ImageNet \cite{deng2009imagenet,russakovsky2015imagenet}. 
From the perspective of caption generation, the Oscar \cite{li2020oscar} single-stream architecture has emerged as an influential model. Oscar
enforces
grounding between image-caption pairs by using object labels as anchor points \cite[a strategy also adopted by][]{hu2020vivo}.  This makes it particularly suited to the goals of this paper, namely, in-depth analysis of the cross-modal interactions in the treatment of objects during generation. Oscar and its successors, VinVL \cite{zhang2021vinvl} and {\sc lemon} \cite{Hu2021} achieved SOTA performance on captioning tasks such as COCO and \texttt{nocaps}.

\paragraph{Methods}
In this paper, we focus on three techniques for model analysis: 
attention analysis, multimodal ablation and probing. 
%
Analyses of attention in pre-trained \vl models include both quantitative methods  \cite[e.g.][]{abnar2020quantifying} and qualitative analysis 
\cite[e.g.][]{li2019visualbert,wei2021integrating}. We use both methods to study how VinVL deploys attention during the generation, of object-centric, versus scene-level captions (Section~\ref{sec:attention}).

Several methods have been proposed to study the extent to which \vl models exploit both visual and textual information
\cite{Shekhar2017,Parcalabescu2022,Gat2021,Hessel2020}. 
Ablation methods 
analyse model behaviour when portions of the input are masked or deleted \cite{bugliarello2021multimodal,cafagna2021vision}. 
We use the ablation of diagnostic objects in scenes (Section \ref{sec:ablation}), to study the reliance of VinVL on such objects during scene-level caption generation.

Probes are well-suited to test for the presence of task-relevant information in model representations
 \cite{Belinkov2019,Belinkov2022}.  \citet{cao2020behind} develop a probe-based benchmark centred around different \vl tasks. \citet{Salin2022} analyse models' reliance on text versus vision to capture colour information. 
\citet{hendricks2021probing} rely on probes to study lexical and syntactic
understanding in \vl models.
In our approach, similar in spirit, we develop probes to identify and measure the extend to which scene information is present in the model's representations before and after fine-tuning on scene-level caption generation.


\section{Data}
\label{sec:data}
We developed the new High Level Scenes (HL-scenes) dataset, which is explicitly designed to pair images with both object-centric and scene-level captions. To this end, we sampled 15k images from the 2014 COCO train split \cite{Chen2015}, with the constraint that each image depicts at least one person. Captions in COCO are highly object-centric \cite{lin2014microsoft}.
We crowd-sourced three scene-level annotations per image on Amazon Mechanical Turk\footnote{Workers were paid at the rate of €0.03 per item, an amount we consider equitable for the work involved, and in line with rates for similar tasks.}, from workers with at least an 85\% approval rating. Crowd workers saw an image and wrote a description in response to the question: \textit{Where is
the picture taken?} Annotators were encouraged to use their knowledge of typical scenes in writing their descriptions.
Finally, we paired our scene-level HL captions with the previously available COCO \cite{lin2014microsoft} captions. 

\begin{figure}[!t]
    \centering
    \includegraphics[scale=0.33]{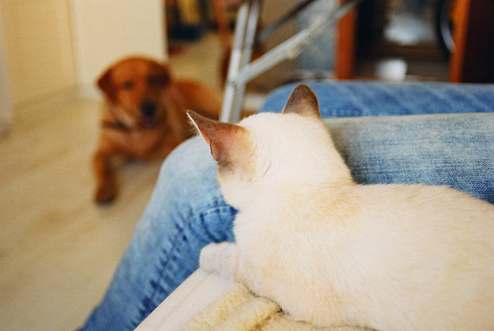}
    \begin{minipage}[b]{\linewidth}
        \footnotesize
         \vspace{\baselineskip}
        {\bf COCO} \\
        {\it Reference:} a close-up of a kitten looking at a dog laying in the background.\\
        {\it Generated:} a cat and a dog sitting next to each other.\\ \\
        {\bf  HL-scenes} \\
        {\it Reference:} in the home. \\
        {\it Generated:} the picture is taken in a house. \\
    \end{minipage}
    \caption{Scene-level captions in HL-Scenes, with corresponding object-centric COCO caption. The generated captions are outputs from VinVL before and after fine-tuning (see Section~\ref{sec:model}).}
    \label{fig:hl-coco-example}
\end{figure}

Figure \ref{fig:hl-coco-example} shows an example of an image with the two types of captions. See Appendix~\ref{sec:app-hl} for more examples. We collected a total of  14,997 image-caption pairs, and we reserve 11,999 for training and 1,499 each for validation and testing.

\section{Model}
\label{sec:model}


VinVL \cite{zhang2021vinvl} is a single-stream BERT-based model with a Faster-RCNN \cite{ren2015faster} visual backbone. It is an extension of 
Oscar \cite{li2020oscar}.
VinVL implements a training strategy where object tags are used as anchor points between the visual and textual modality to facilitate cross-modal alignment. 
As pointed out by \citet{li2020oscar}, this strategy is motivated by the fact that in the datasets used to pre-train multimodal models, between 1 and 3 of the objects detected by the visual backbone are mentioned in the caption. However, the object labels are provided by an off-the-self object detector separately trained on Visual Genome 
\cite{krishna2017visual}.
VinVL was pre-trained on a combination of COCO \cite{Chen2015}, Conceptual Captions \cite{Sharma2018}, SBU captions \cite{Ordonez2011} and Flickr30k \cite{young-etal-2014-image}, as well as additional VQA data.

VinVL has been shown to perform well on understanding tasks, including
VQA, NLVR2, image-text and text-image retrieval \cite{goyal2017making, Suhr2019, lin2014microsoft}, and on generative tasks, including COCO \cite{Chen2015} and \texttt{nocaps} \cite{agrawal2019nocaps}. 

In the Oscar family of models, the use of labels as anchors makes the models ideal for our experiments, in that it explicitly enables us to study the interaction between object-level information (captured by labels and visual features) and scene-level description generation.

\subsection{Fine-tuning}
We first establish that VinVL can generate scene descriptions after fine-tunning, before turning to an in-depth analysis of 
the model's attention and internal representations. 

We note that since the HL-scenes dataset extends the COCO dataset, the model has been exposed to the images of the HL-scenes dataset during pre-training on COCO. On the other hand, the scene descriptions are completely novel. 

We fine-tune on scene-descriptions for 10 epochs. We use the standard configuration used by \citet{zhang2021vinvl} for image captioning. At inference time, we fix the maximum generation length to 20 tokens and use a beam size of 5. 

VinVL shows a quick adaptation to the scene-level descriptions from the first epoch. This adaptability recalls observations made for other
transformer-based generative models \cite[e.g.][]{brown2020language}. 
We show an example in Figure~\ref{fig:hl-coco-example}. For completeness, Table~\ref{metrics_hl} reports the automatic evaluation metrics computed on the validation set over 10 epochs. For more details see Appendix~\ref{sec:ft_details}.

\begin{table}[t]
\small
\begin{tabular}{l|cccccc}
    \hline
    Epoch. & B4 & M & RL & CIDEr & SPICE \\ \hline
    2 & 49.3    & 29.3   & 67.1     & 161.8 & 32.6  \\ 
    4  & 49.7    & 30.1   & 68.1     & 168.5 & 34.0  \\ 
    6  & 48.5    & 29.8   & 67.3     & 164.9 & 33.5  \\ 
    8  & 48.9    & 30.2   & 67.6     & 165.8 & 33.9  \\ 
    10  & 49.1    & 30.4   & 67.7     & 168.0 & 34.4  \\ \hline
    \end{tabular}
    \caption{\label{metrics_hl} Automatic metrics computed over different epochs on the HL-Scenes validation set. B4: Bleu-4; M: METEOR; RL: ROUGE-L.}
\end{table}

\section{How does attention to objects change from object-centric to scene-level generation?}\label{sec:attention}
We first investigate the model's self-attention before and after fine-tuning on the scene-level caption generation task.


\paragraph{Method}
We focus on the self-attention patterns in the first layer, as they are directly connected to the inputs and do not depend on higher-level interactions which might obscure the fundamental changes in attention across the two modalities (visual features and labels) in VinVL. A discussion of attention patterns at higher layers can be found in Appendix (\ref{sec:att_details}).
We select 100 random samples from the HL-Scenes test-set and extract the attention matrices before and after fine-tuning on scene descriptions. We aggregate the attention values by taking the maximum across all the heads, as it allows us to observe where the model tends to assign a significant amount of attention, giving us a better view of the potential impact of fine-tuning on scene-level captions. 
VinVL prevents textual inputs from directly interacting with the other modalities during generation; therefore there is no interaction between caption tokens and visual features. On the other hand, the model includes object tags as anchors and this allows us to study the multimodal interactions between the visual features and these object labels.


\begin{figure*}[]
    \centering
    \begin{subfigure}[b]{0.45\linewidth}
        \includegraphics[scale=0.15]{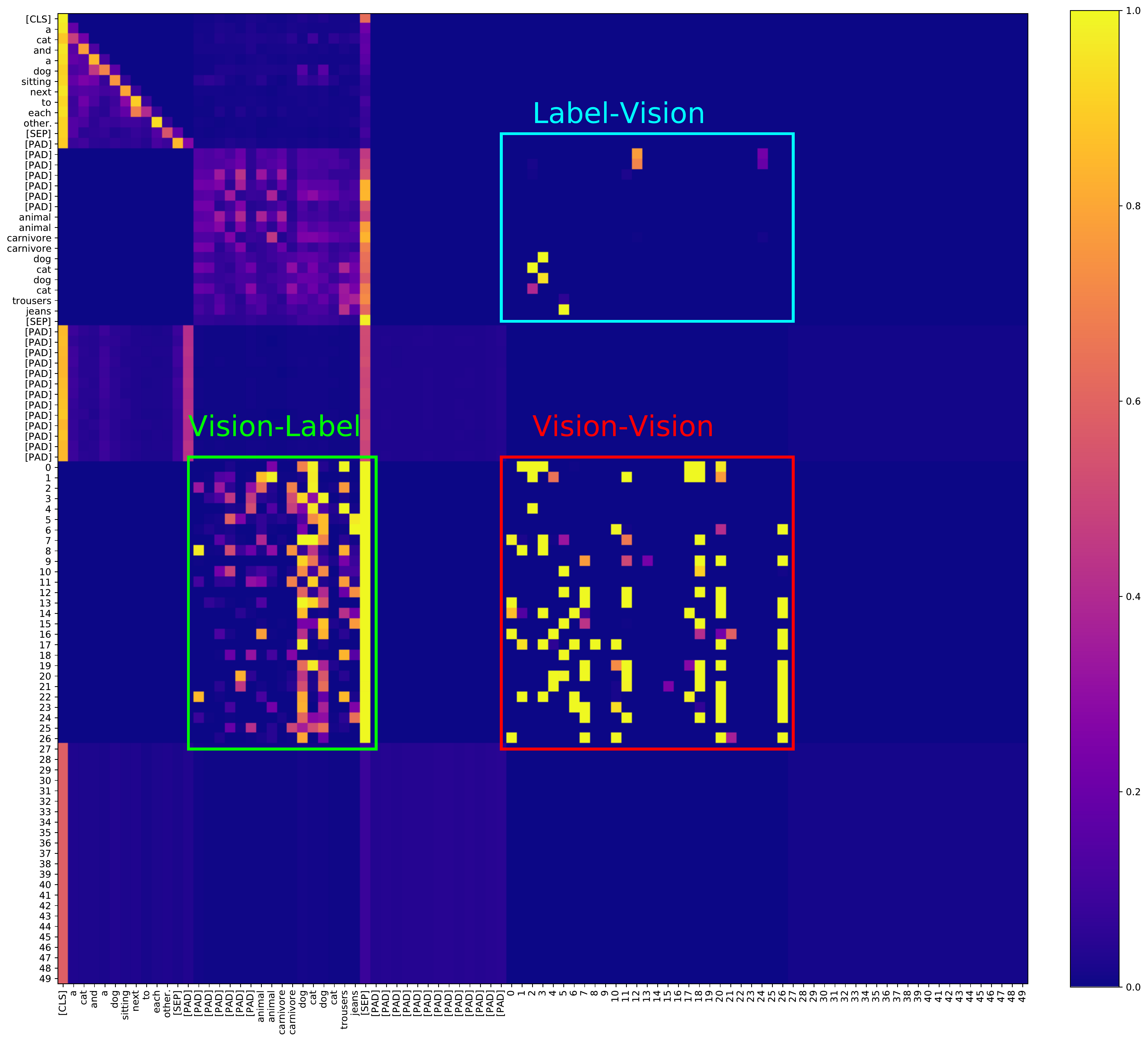}
        \caption{Attention matrix of the pre-trained model}\label{fig:hl-coco-ex-pre}
    \end{subfigure}
     \hfill
    \begin{subfigure}[b]{0.45\linewidth}
        \includegraphics[scale=0.15]{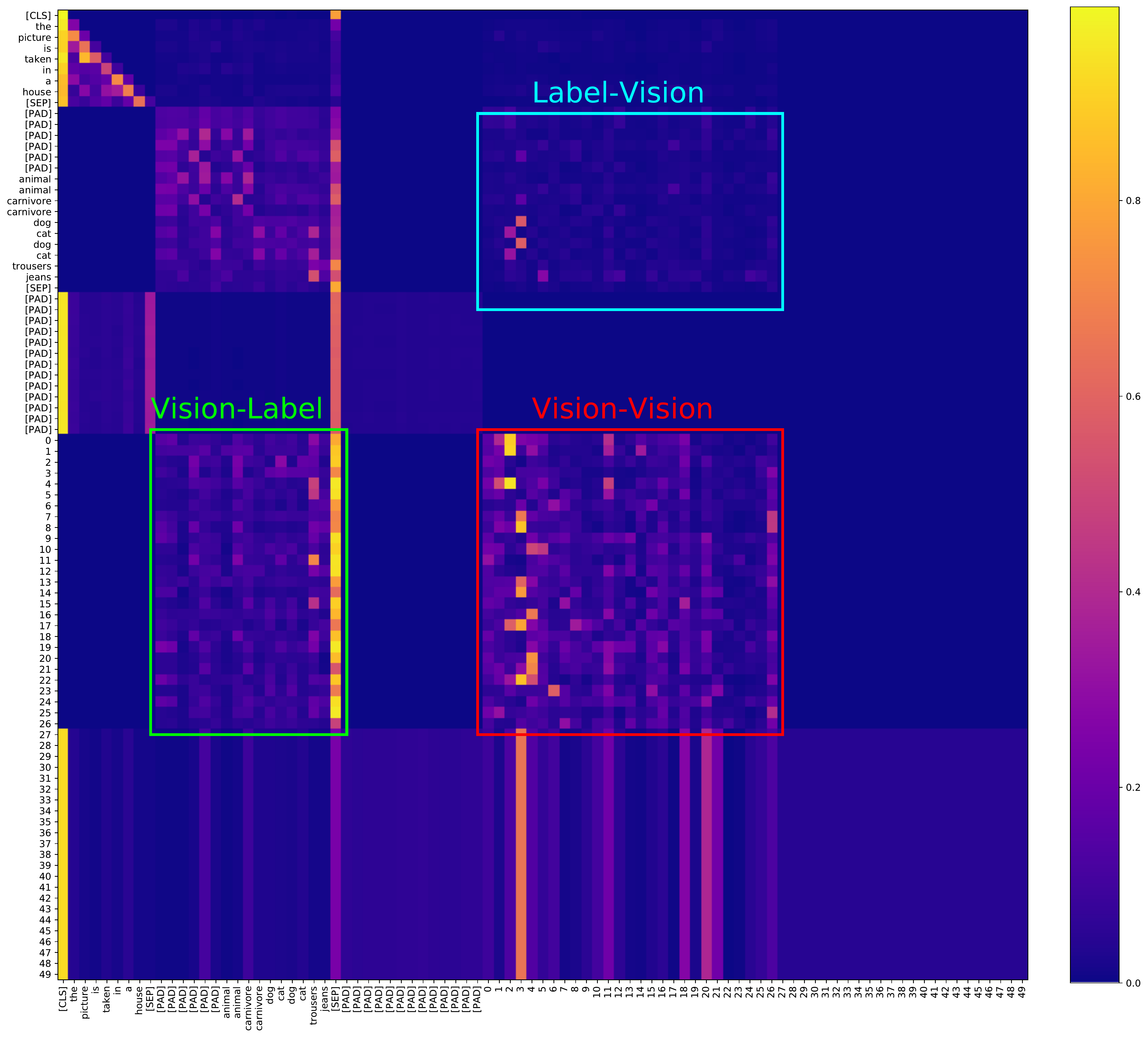}
        \caption{Attention matrix of the fine-tuned model}\label{fig:hl-coco-ex-ft}
    \end{subfigure}
    \caption{Attention matrices comparison for the image in Figure~\ref{fig:hl-coco-example}. We highlight the sub-blocks corresponding to \textcolor{red}{vision-to-vision}, \textcolor{green}{vision-to-label} and \textcolor{cyan}{label-to-vision}. In the pre-trained model, attention mass is sharply focused on individual portions of the input; after fine-tuning, a more even distribution is observed.}
    \label{fig:attention-comparison}
\end{figure*}

\paragraph{VinVL acquires a holistic view of the scene after pre-training}
Figure \ref{fig:attention-comparison} is a representative example of self-attention matrices extracted from the pre-trained (\ref{fig:hl-coco-ex-pre}) and fine-tuned (\ref{fig:hl-coco-ex-ft}) model with the image in Figure \ref{fig:hl-coco-example}. The pre-trained model, which generates an object-centric caption, focuses attention on individual input tokens in the \textcolor{red}{vision-to-vision}, \textcolor{green}{vision-to-label} and \textcolor{cyan}{label-to-vision} sub-blocks.

After fine-tuning, as the model generates a scene-level caption, the self-attention 
appears to be more evenly distributed over the inputs (\ref{fig:hl-coco-ex-ft}). 
This suggests that when generating scene-level captions, the model leverages a wider range of visual features with less exclusive focus on individual objects or labels.

\begin{figure}[!h]
    \centering
    \includegraphics[scale=0.49]{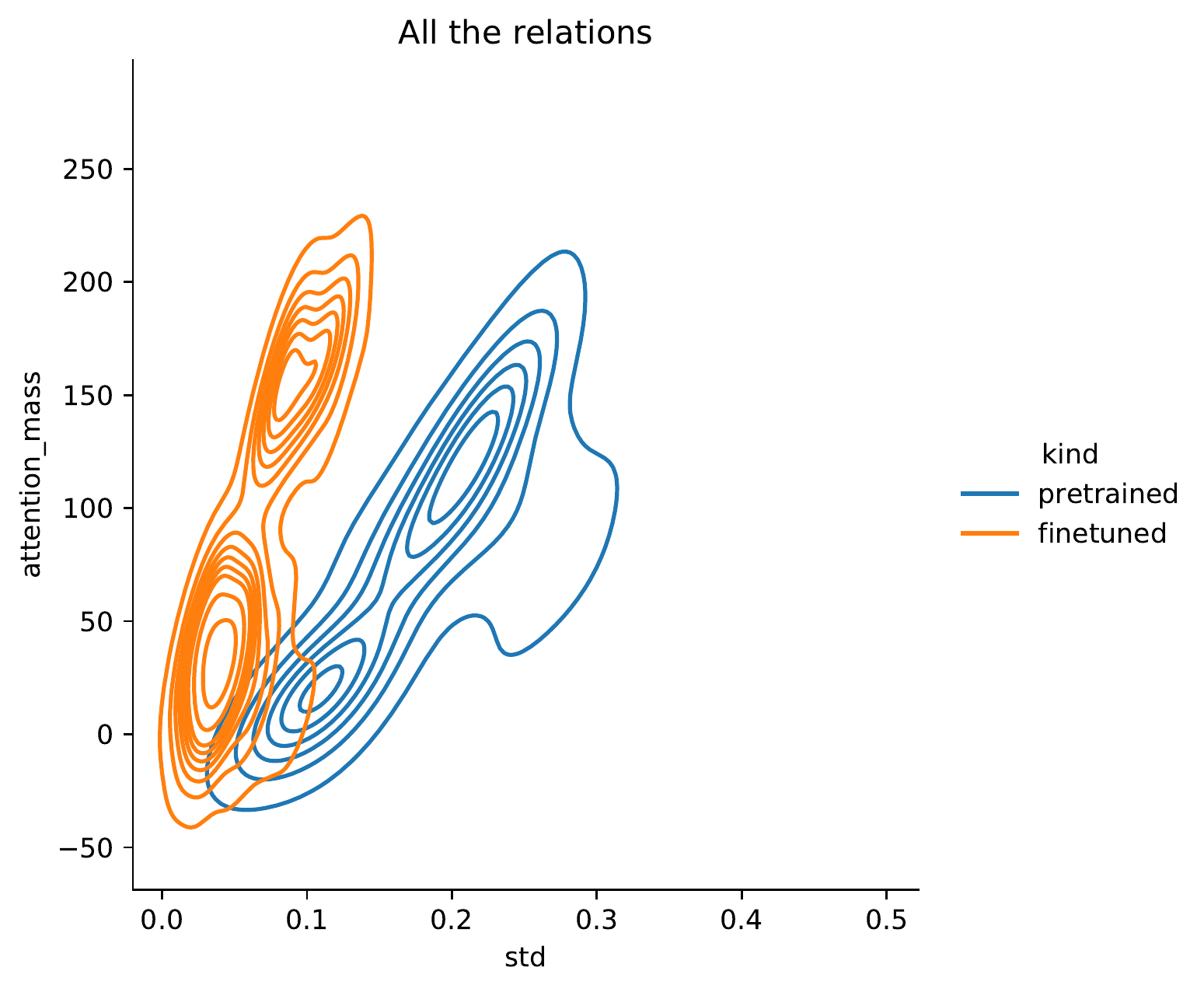}
    \caption{Kernel density estimate of distributions of standard deviations against attention mass for pre-trained and fine-tuned VinVL.}
    \label{fig:att-std}
\end{figure}

We perform a quantitative analysis of the self-attention in the sub-blocks of the matrix
involving visual regions and object labels, computing a
kernel density estimate of the distributions of the standard deviations and attention masses 
for each of the 100 samples. The result is shown in Figure~\ref{fig:att-std}. 
It is clear that the fine-tuned model has overall a lower standard deviation than the pre-trained model. This confirms that
a similar attention mass is distributed more evenly after fine-tuning.
We take this as evidence that in the process of generating scene descriptions, the fine-tuned model acquires a more holistic view of the input image, in contrast to the highly object-centred deployment of attentional resources evident in the pre-trained model.


\paragraph{VinVL relies on diagnostic objects when generating scene-level captions}
VinVL redistributes self-attention over a wider range of visual features after fine-tuning. Nevertheless, previous work on scene perception \cite{Self2019,Vo2021} leads us to expect that in describing a scene, the model needs to rely on highly diagnostic objects. We compute diagnosticity empirically, based on the occurrence of objects in scenes in our dataset. 
Let $S$ be the set of the $k$ most frequent scene types mentioned in scene-level captions in the HL-Scenes dataset.\footnote{Since our dataset consists of captions, we extract scene labels from these captions. See Appendix (\ref{sec:att_details}).} We proceed as follows:


\begin{enumerate}
    \item $ \forall \; s \; \in \; S$ we build $ O_M^s = [ o_1^s, o_2^s, ..., o_n^s ] $, 
    the ranked list of the $n$ most attended objects by the model $M$ when generating a description of a scene of type $s$. 
    \item Similarly, $ \forall \; s \; \in \; S$ we collect $ O_D^s = [ o_1^s, o_2^s, ..., o_n^s ] $, the ranked list of the most frequent objects in images depicting scenes of type $s$ in the dataset $D$. 
\end{enumerate}
We measure the overlap between $O_M^s$ and $O_D^s$
by computing their Intersection over Union (IoU), which is only sensitive to overlap in content, as well as 
their Rank Biased Overlap \cite[RBO;][]{webber2010similarity}\footnote{\href{https://github.com/changyaochen/rbo}{https://github.com/changyaochen/rbo}}, which computes the similarity of two ranked lists. More details about this metric are given in Appendix~\ref{sec:att_details}. 
\begin{table}[t!]
\small
\centering
\begin{tabular}{l|lll|lll}
\hline
\multirow{2}{*}{\textbf{Scene}} &
  \multicolumn{3}{c|}{\textbf{RBO @}} &
  \multicolumn{3}{c}{\textbf{IoU @}} \\ \cline{2-7} 
\multicolumn{1}{c}{} &
  \multicolumn{1}{|c}{\textbf{3}} & 
  \multicolumn{1}{c}{\textbf{5}} &
  \multicolumn{1}{c|}{\textbf{7}} &
  \multicolumn{1}{c}{\textbf{3}} &
  \multicolumn{1}{c}{\textbf{5}} &
  \multicolumn{1}{c}{\textbf{7}} \\ \hline
station &
  \multicolumn{1}{l}{0.88} &
  \multicolumn{1}{l}{0.84} &
  0.87 &
  \multicolumn{1}{l}{0.5} &
  \multicolumn{1}{l}{0.66} &
  1.0 \\ 
road &
  \multicolumn{1}{l}{1.0} &
  \multicolumn{1}{l}{0.9} &
  0.91 &
  \multicolumn{1}{l}{1.0} &
  \multicolumn{1}{l}{0.66} &
  1.0 \\ 
room &
  \multicolumn{1}{l}{0.27} &
  \multicolumn{1}{l}{0.25} &
  0.24 &
  \multicolumn{1}{l}{0.2} &
  \multicolumn{1}{l}{0.11} &
  0.18 \\ 
sea &
  \multicolumn{1}{l}{0.88} &
  \multicolumn{1}{l}{0.84} &
  0.8 &
  \multicolumn{1}{l}{0.5} &
  \multicolumn{1}{l}{0.66} &
  0.55 \\
resort &
  \multicolumn{1}{l}{0.72} &
  \multicolumn{1}{l}{0.7} &
  0.7 &
  \multicolumn{1}{l}{0.5} &
  \multicolumn{1}{l}{0.42} &
  0.55 \\ 
house &
  \multicolumn{1}{l}{0.38} &
  \multicolumn{1}{l}{0.5} &
  0.53 &
  \multicolumn{1}{l}{0.5} &
  \multicolumn{1}{l}{0.42} &
  0.55 \\ 
restaurant &
  \multicolumn{1}{l}{0.55} &
  \multicolumn{1}{l}{0.55} &
  0.54 &
  \multicolumn{1}{l}{0.5} &
  \multicolumn{1}{l}{0.42} &
  0.53 \\ \hline
\end{tabular}
\caption{\label{rbo} Rank Biased Overlap (RBO) and Intersection over Union (IoU) of the most attended objects and the most frequent objects for the top seven common scenes. Both metrics range from 0 (no overlap) to 1 (perfect correspondence).}
\end{table}
Table~\ref{rbo} shows RBO and IoU for the top 3, 5 and 7 objects in the lists.
We observe that the two metrics correlate strongly ($r(19)=.81, p<.001$). From this we conclude that during generation of scene-level captions, the model attends more to diagnostic objects, i.e. those which
are common in a scene of a given type.
Moreover, we observe high scores for scene types such as \textit{station, road, resort, sea}. In our dataset, these are characterised by frequently occurring objects, which are therefore highly diagnostic of scene type.
In contrast, for scenes like \textit{room, house, restaurant} we observe lower scores. We hypothesise that this is due to the fact that such scenes can contain a wider variety of objects, which individually have lower diagnosticity with respect to the scene type.


\section{How reliant is the model on diagnostic objects?}
\label{sec:ablation}
The results from the previous sections established that, following fine-tuning on scene-level descriptions, VinVL distributes attention more evenly over objects in a scene. Nevertheless, the objects which are most likely to be present in a scene attract the highest proportion of the attention mass. This raises the question whether, by removing highly diagnostic objects from an image, the model representations are still informative enough to detect what type of scene is represented in an image. 

We first address this issue from the perspective of generation: does a model fine-tuned on scene descriptions still manage to correctly describe a picture at the scene level, when highly diagnostic objects are unavailable? Given the more even distribution of attention observed across scene components in the fine-tuned model, our hypothesis would be that even in the absence of such highly diagnostic objects, the model can rely on other information to detect the scene type. Hence, we expect the fine-tuned model to be more robust to object ablation in the visual modality, compared to the model pre-trained on object-level captions. 


\subsection{Method}
As explained in Section~\ref{sec:model}, in VinVL, two separate models are used to (i) extract visual features corresponding to regions via the model's visual backbone; and (ii) to determine the object labels that function as anchors between the visual and textual modalities.
This means we do not have an exact correspondence between object labels and visual features.

\paragraph{Visual feature tagging}
For simplicity we will refer to \textit{vf} as the bounding box a visual feature corresponds to, and \textit{ot} as the bounding box an object label corresponds to. To perform an ablation, we first establish an approximate correspondence betweeen \textit{ot} and \textit{vf}, using \textit{ot} as reference to assign an object label to the visual features.

We compute the IoU\footnote{Note that in this section we refer to the Intersection Over Union to compute the overlap between two bounding boxes, not the metric used to compute the overlap between two sets of items as done in Section~\ref{sec:attention}.}  between \textit{vf} and \textit{ot} and empirically assign a label to a visual feature if $IoU(vf, ot) >=0.6$. 
Moreover, if \textit{vf} is contained by or overlaps with \textit{ot} by at least 80\% of its area, we assign to \textit{vf} the label of \textit{ot}. 
With this heuristic we cover 74\% of the visual features of every image of our sample.

\paragraph{Computing object diagnosticity}
We use the scene labels extracted from captions in Section~\ref{sec:attention}, and compute the Pointwise Mutual Information (PMI)
between scene types and object labels. Examples of the most informative objects for some scenes are shown in Table~\ref{informative_objects}.

\begin{table}[]
\small
\begin{tabular}{l|l}
    \hline
    Scene & Top informative objects \\ \hline \hline
    restaurant & french fries, fork, submarine sandwich \\
    road & vehicle number plate, traffic sign, traffic light \\
    sea & surfboard, watercraft, boat \\
    room & computer mouse, nightstand, tablet computer \\
    station & train, suitcase, luggage and bags \\ \hline
    \end{tabular}
    \caption{\label{informative_objects} Most informative objects for some scenes ranked using PMI.}
\end{table}

\paragraph{Ablation}
Ablation of an object is performed similarly to
\cite{frank2021vision}, by removing its corresponding label from the list of object tags, along with every visual feature assigned to that object. We replace them with a \texttt{[PAD]} token. We compare captions generated by both the pre-trained and fine-tuned model with and without ablation of the top 1, 2 and 3 most informative objects for a given scene in the test-set. For more details on the sample sizes see Appendix~\ref{sec:app-ablation}.

\subsection{Results}
We expect to observe some differences in the generations when ablation is applied, especially in the pre-trained model, as the ablation removes information which is explicitly verbalised in object-centric captions. For the pre-trained model, object-centric captions change 41\% of the time after ablation, compared to 13\% of the time for the scene-level captions by the fine-tuned model. 


A manual inspection on a sample of items suggested that the changes in the captions involve minimal semantic shifts, often due to minor function word changes or a more generic term being generated for the noun denoting the scene type. Some examples are shown in Figure \ref{fig:abl-ex}.

\begin{figure}
    \centering
    \small
    \fbox{\begin{minipage}[b]{2.9in}
        the picture is \textcolor{green}{shot in a ski resort} $\rightarrow$ the picture is  \textcolor{red}{taken in a snowfield} \textit{(jacket, tree, footwear)}
        
        the picture is \textcolor{green}{shot in a baseball field}  $\rightarrow$ the picture is \textcolor{red}{taken in a ground} \textit{(sports uniform, man, boy)}

        in \textcolor{green}{a} kitchen $\rightarrow$ in \textcolor{red}{the} kitchen \textit{(kitchen appliance, countertop, cabinetry)}
    \end{minipage}}

    \caption{Changes to scene-level captions generated by the fine-tuned model after ablation of three diagnostic objects. Ablated objects are shown in parentheses.}
    \label{fig:abl-ex}
\end{figure}


In summary, the model is resilient to ablation in the visual modality, suggesting that its representations are robust for both types of generation task, but more so for scene-level captioning. 
We study robustness of representations in more detail using probes, in Section~\ref{sec:prob}.


\begin{figure}[!t]
    \includegraphics[scale=0.48]{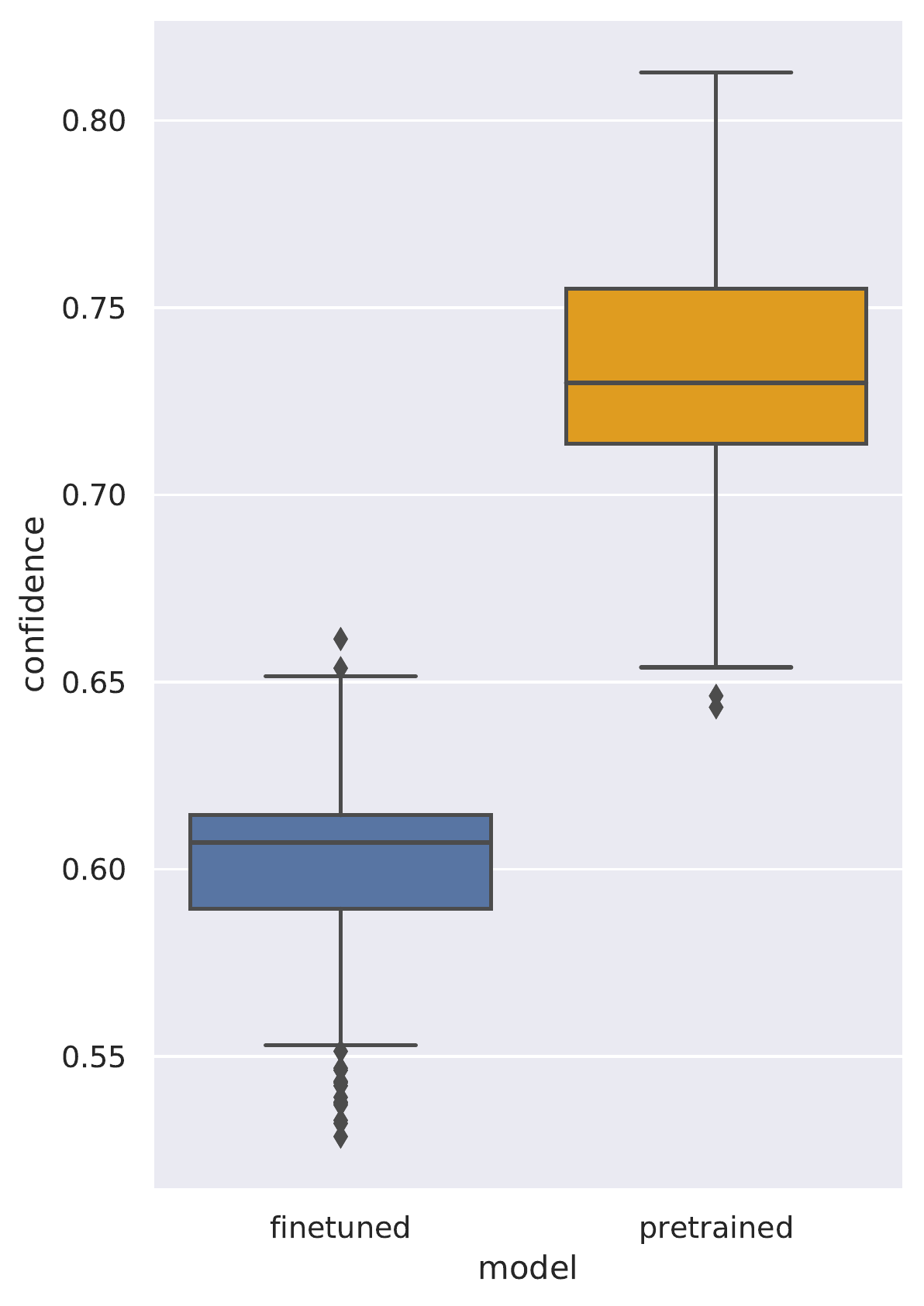}
    \caption{Confidence scores of the unchanged caption after ablation. On the left, the model generating scene-level descriptions (fine-tuned); on the right, the model generating objective descriptions (pre-trained). }
    \label{fig:conf}
\end{figure}

\begin{figure}[!t]
    \includegraphics[scale=0.53]{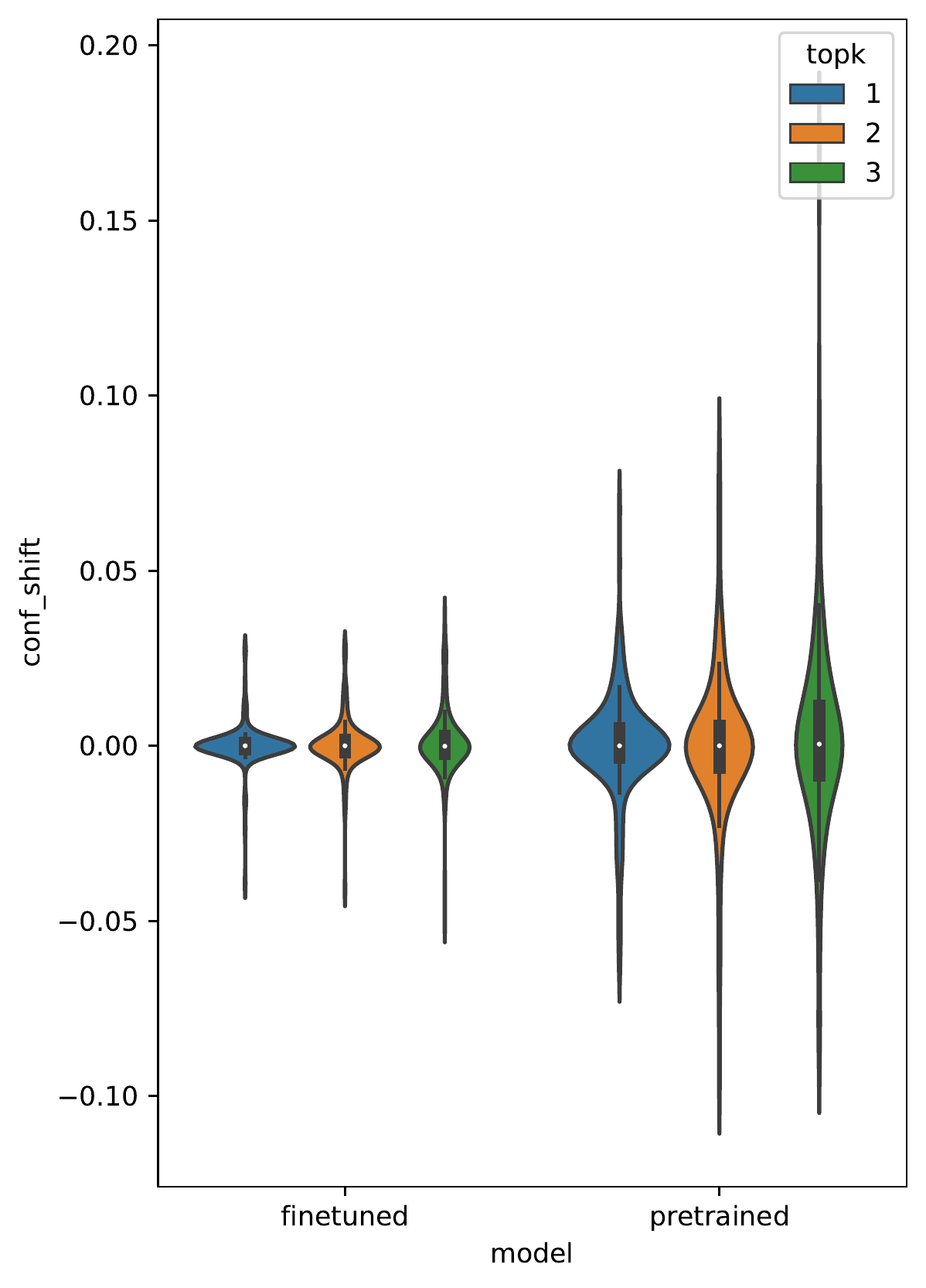}
    \caption{Confidence shift of the unchanged captions when ablating the top 1, 2 and 3 most informative objects from the scene. A negative shift means that the caption was generated with higher confidence after ablation.
    On the left, the model generating scene-descriptions (fine-tuned); on the right, the model generating object-centric descriptions (pre-trained). }
    \label{fig:violin}
\end{figure}

\paragraph{Confidence scores}  
We also analyse the confidence score produced at generation time by the model for those captions which do not change after ablation.
As shown in Figure~\ref{fig:conf}, after ablation pre-trained VinVL generates object-centric descriptions with higher confidence than fine-tuned VinVL does with scene-level descriptions. However, the variance in the confidence score after ablation is lower for the fine-tuned model generating scene-level captions (Figure~\ref{fig:violin}), suggesting greater robustness to ablation during scene-level caption generation.

\section{Can we disentangle the role of attention and model representation?}
\label{sec:prob}
The results so far suggest that there are significant changes in the model's self-attention, though it relies on diagnostic objects to generate scene-level captions. It is also somewhat more robust to object ablation, especially in the fine-tuned case. At this point, we probe the model's representations to address to what extent the knowledge required for scene-level caption generation is already present after pre-training. This would imply that the primary change to the model after fine-tuning is in the self-attention mechanism. 

\begin{figure}[!t]
    \includegraphics[scale=0.48]{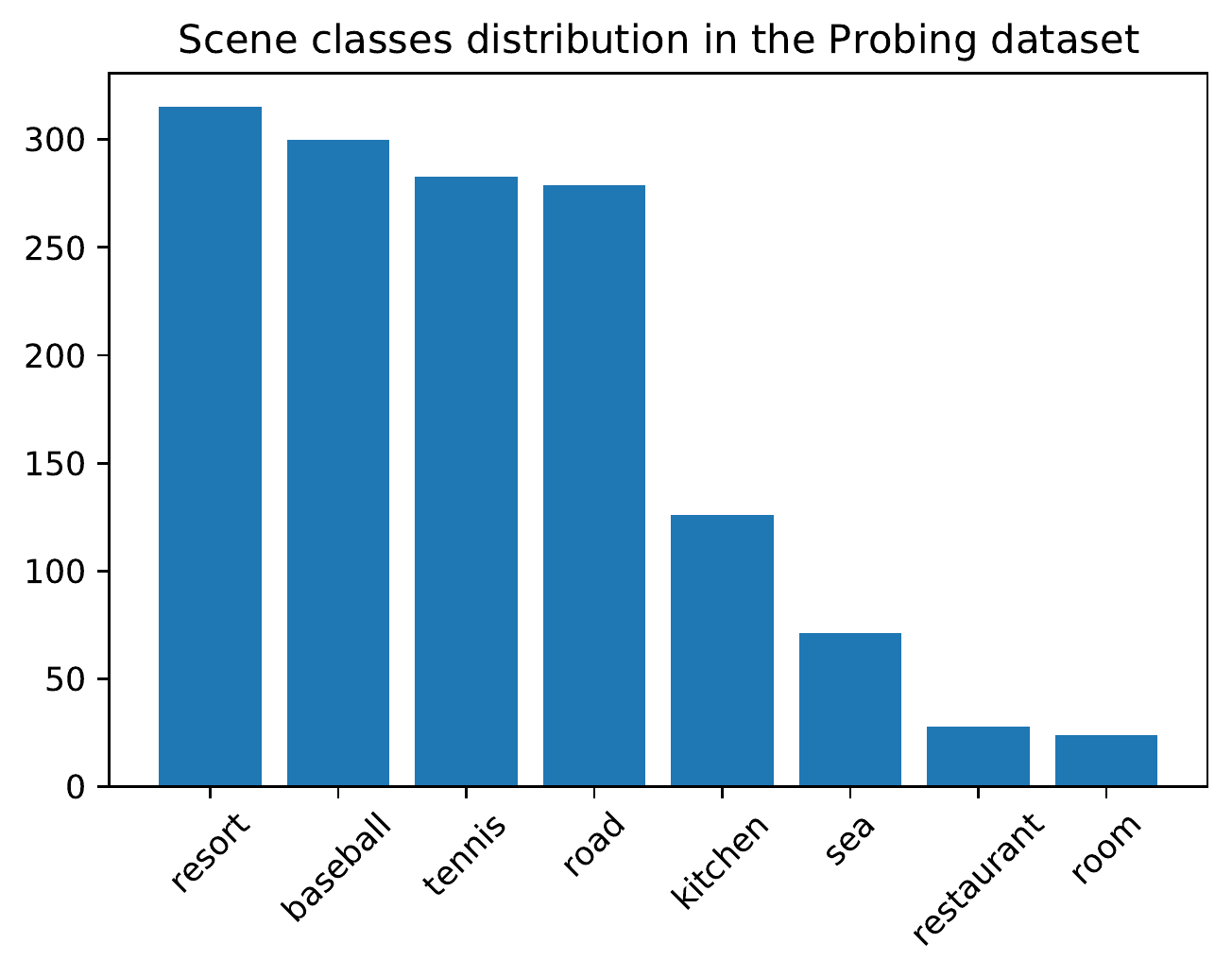}
    \caption{Scene distribution in the probing dataset}
    \label{fig:class_dist}
\end{figure}

\paragraph{Method}
Given a pair $(V, L)$ consisting of visual features $V$ and object labels $L$, we train a probe to classify scene type based on VinVL encodings, before and after fine-tuning. We also repeat the procedure on inputs ablated as described in
Section~\ref{sec:ablation}.
For this experiment, we identify 1426 images from HL-scenes, representing 8 types of scene, downsampling the more frequent classes (see Appendix~\ref{sec:prob_details} for details). The class distribution is shown in Figure~\ref{fig:class_dist}.
For every image in the probing dataset we extract the model's feature representations from the last layer and we average across the inputs, obtaining a single vector.


We train both a neural and a random forest probe. We report results from the latter which is the best performing; full details of the neural probe are in Appendix \ref{sec:prob_details}. 

\begin{figure}
\centering
    \includegraphics[scale=0.42]{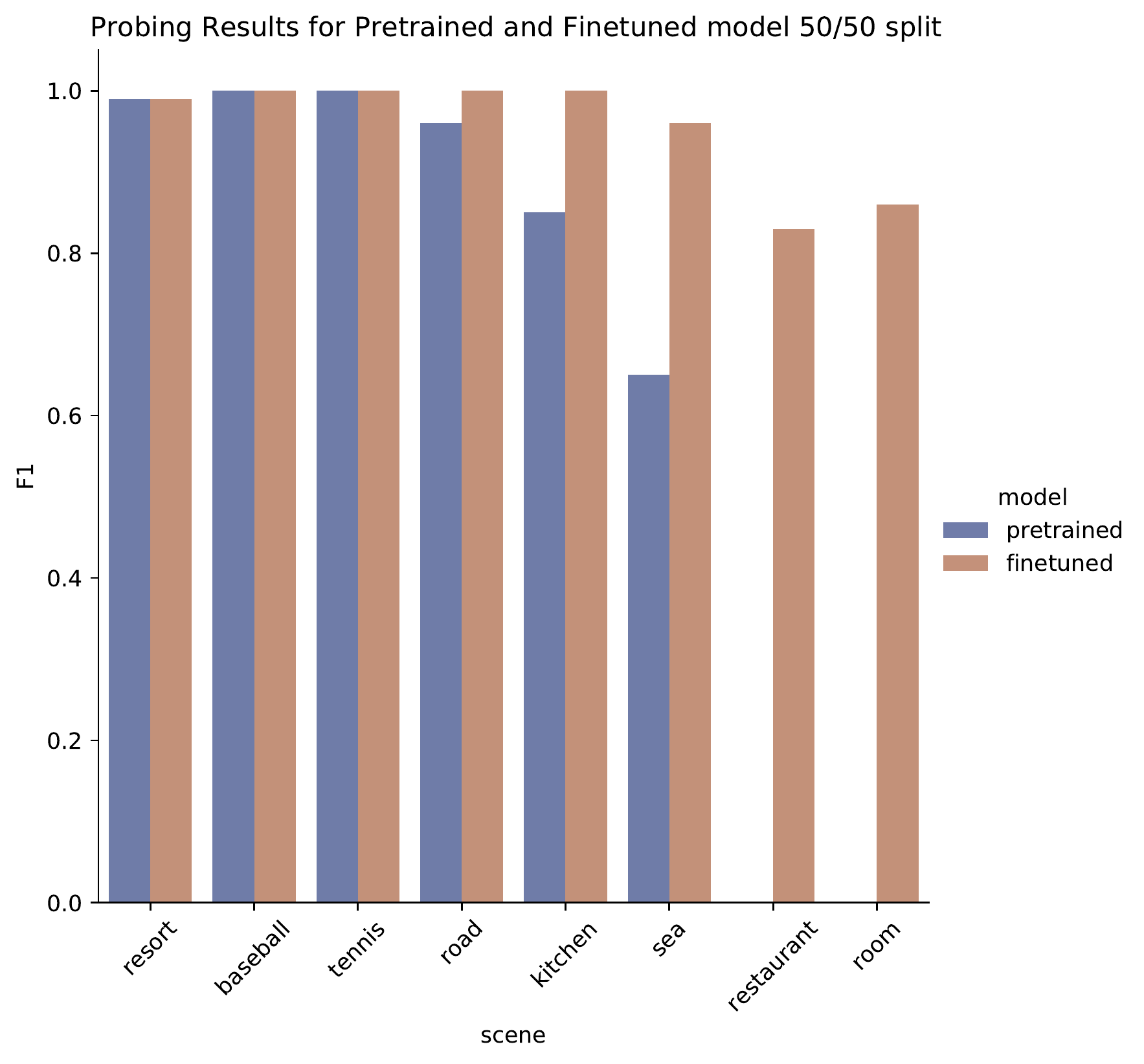}
    \caption{F1-scores of the scene classification task for the pre-trained in (blue) and the fine-tuned model (orange).}
    \label{fig:probe_ft_pre}
\end{figure}

\paragraph{Results}
Probes are tested on different train/test proportions, up to a 50/50 split. In Figure~\ref{fig:probe_ft_pre} we report results for the 50/50 train/test split, which is also the most challenging (for results on other splits see Appendix~\ref{sec:prob_details}). The baseline performs a random assignment of the labels to the features. For both pre-trained and fine-tuned models, probes perform at ceiling for scenes with a high support (cf. Figure~\ref{fig:class_dist}). For scene types with a very low frequency, like \textit{restaurant} and \textit{room}, the probe trained on features from the pre-trained model fails. 
In contrast, probing features from the fine-tuned model still performs at ceiling. These results suggest that the information to detect the scene type is already present to some extent in the pre-trained model. Nevertheless, fine-tuning proves effective in closing the gap for low-support scenes.  

When trained on features extracted from ablated inputs in Table~\ref{probe_f1}, the probe is not particularly affected by the ablation, confirming the robustness of the model's representations as observed in the ablation study (Section~\ref{sec:ablation}).



\begin{table}[t]
\small
\begin{tabular}{l|ccc}
\hline
Model                 & micro-F1 & macro-F1 & weighted-F1 \\ \hline \hline
Random                & 0.16     & 0.12     & 0.16        \\ \hline 
Pretrained            & 0.94     & 0.67     & 0.92        \\ 
Finetuned             & \bf{0.99}     & \bf{0.96}     & \bf{0.99}        \\ \hline
Pretrained (A) & 0.92     & 0.66     & 0.90        \\ 
Finetuned (A)  & 0.98     & 0.88     & 0.97        \\ \hline

\end{tabular}
\caption{\label{probe_f1} F1-scores for the scene classification task in the 50/50 split using a random forest. The first row (Random) corresponds to the performance of random baseline while (A) is the performance on the features obtained by the ablating the input.}
\end{table}

\section{Conclusion}
\label{sec:conclusion}
In this paper, we addressed scene-level caption generation. Taking a cue from prior work on scene semantics and syntax, our goal was to assess \vl models' ability to reason about the link between scenes and their components and exploit this to generate informative captions with less redundancy.

\paragraph{Findings and Contributions}
We contributed a new dataset pairing object-centric and scene-level captions, and showed that VinVL is able to generate scene-level descriptions with minimal fine-tuning.

Our analysis showed that the fine-tuning results in a more even distribution of attention mass over the image, suggesting a more `holistic' view of the scene which nevertheless makes use of diagnostic object information. Using a combination of ablation and probing methods, we also show that much of the relevant information for scene-level captioning is present after pre-training. Hence, the model's ability to generate scene-level captions is primarily acquired through a change in its self-attention. 

\paragraph{Limitations}
In this work we draw conclusions from an analysis of a single model, this can be considered a limitation. Nevertheless, VinVL is representative of a larger family of SOTA models in the field, based on Oscar, which are dominating the scene in V\&L tasks. Moreover, Oscar pretraining  using object tags makes the model well-suited to an in-depth analysis of cross-modal interactions in a generative context.

We acknowledge also that the results of the ablation analysis (Section~\ref{sec:ablation}) could in part be affected by the approximate nature of our tagging method. Furthermore, as noted by \citet{frank2021vision}, visual feature deletion may still leave relevant contextual information in the remaining feature vectors, due to the Faster-RCNN's wide field of view.


\section*{Acknowledgements}
Contribution from the ITN project NL4XAI (\textit{Natural Language for Explainable AI}). This project has received funding from the European Union’s Horizon 2020 research and innovation programme under the Marie Skłodowska-Curie grant agreement No 860621. This document reflects the views of the author(s) and does not necessarily reflect the views or policy of the European Commission. The REA cannot be held responsible for any use that may be made of the information this document contains.

\bibliography{anthology,custom}

\clearpage

\appendix

\section*{Appendix}

\section{Fine-tuning Details}
\label{sec:ft_details}
We fine-tune the VinVL pre-trained base version\footnote{\url{https://github.com/microsoft/Oscar/blob/master/VinVL_MODEL_ZOO.md\#Oscarplus-pretraining}} using the original configuration for 10 epochs on scene descriptions. We refer to it as the \textit{fine-tuned} model. Since the HL-scenes dataset images are included in COCO, we use the pre-computed visual features and labels  provided in the original VinVL implementation.

We refer to the \textit{pre-trained} model, as the base model trained on the image captioning task on COCO captions optimized using cross-entropy. All the experiments involving the pre-trained model are performed using the original configuration used in \citet{li2020oscar}.
The fine-tuning is carried out with batch size 32 on a NVIDIA GTX 2080 TI 11 GB.

\section{Self-attention Details}
\label{sec:att_details}
\begin{figure}[t]
\centering
    \includegraphics[scale=0.5]{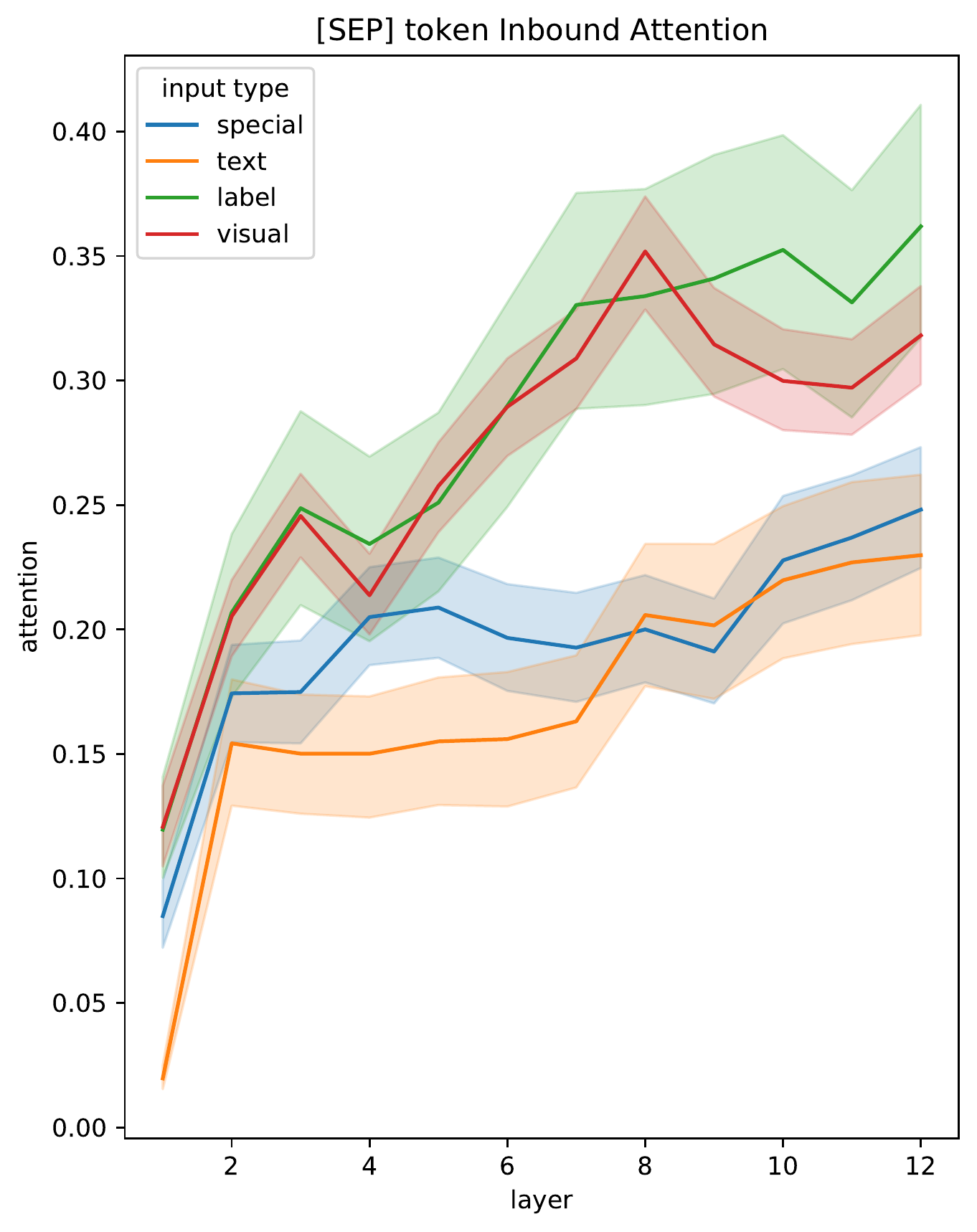}
    \caption{Inbound attention of the \texttt{[SEP]} per input type token across the layers. Special tokens correspond to \texttt{[CLS]}, \texttt{[PAD]} and \texttt{[SEP]}.}
    \label{fig:sep_att}
\end{figure}

\paragraph{Attention beyond Layer 1}
At higher layers the attention converges on the special token \texttt{[SEP]}, used to separate the \textit{text + object tags} from the \textit{visual} input, as shown in Figure~\ref{fig:sep_att}. A similar behaviour has been observed analysing BERT's attention \cite{clark2019does}. 

\begin{figure}[h!]
    \centering
    \begin{subfigure}[b]{\linewidth}
        \includegraphics[scale=0.42]{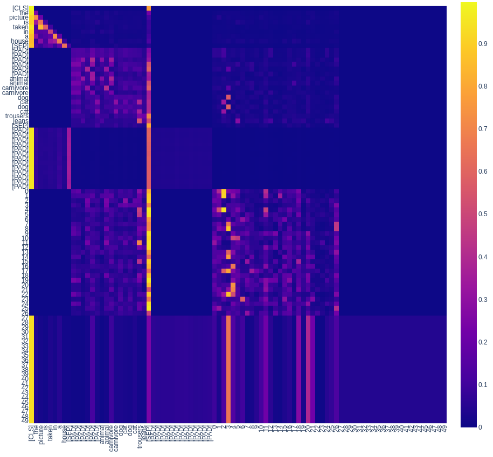}
        \caption{Layer 1}
     \end{subfigure}
     \hfill
    \begin{subfigure}[b]{\linewidth}
        \includegraphics[scale=0.42]{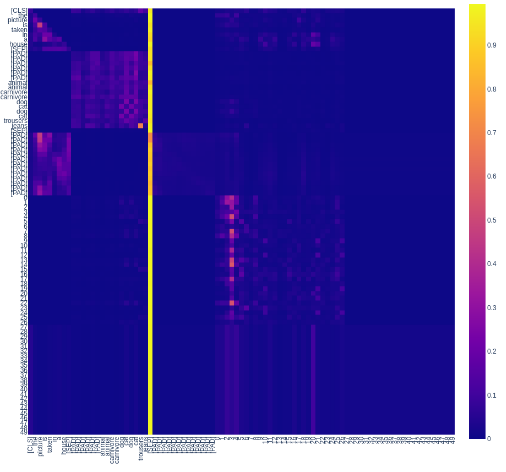}
        \caption{Layer 6}
     \end{subfigure}
     \hfill
    \begin{subfigure}[b]{\linewidth}
        \includegraphics[scale=0.42]{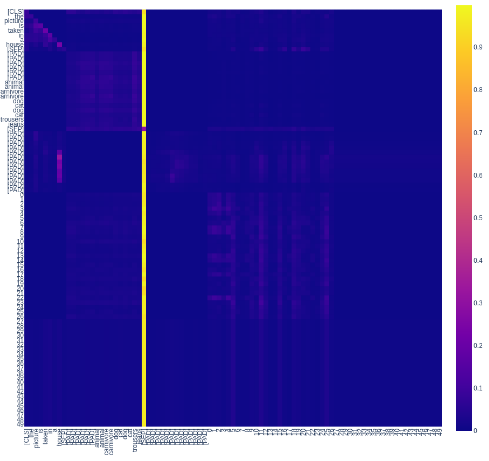}
        \caption{Layer 12}
    \end{subfigure}
    \caption{Attention matrices for layers 1, 6 and 12. The attention weights progressively gather on the \texttt{[SEP]} token.}
    \label{fig:att_layers}
\end{figure}

Figure~\ref{fig:att_layers} shows how this pattern becomes more pronounced as we move further across the layers, preventing from observing any kind of input interplay.
Although the \textit{text}, \textit{object tags} and \textit{visual} sequences can be of different lengths, the \texttt{[SEP]} token sits always in the same position among the inputs, as the padding is always applied to keep the \textit{text + object tags} sequence of the same length. We believe that this regularity is used by the model as a sort of pivot among the inputs. This can cause the a high accumulation of attentional resources by the model. 

\paragraph{Scene label extraction}
As described in Section~\ref{sec:data}, during the data collection, the annotators where asked to answer the direct question: \textit{Where is the picture taken?} As a consequence, the scene-captions often have a regular structure, captured by the following three representative examples:

\begin{itemize}
\item the picture has been taken in a \textit{restaurant}
\item on a \textit{beach}
\item this is in an \textit{airport}
\end{itemize}

To extract the scene labels, we tokenize the scene-captions and we remove punctuation and stop-words (we add \textit{picture} to the list of the standard stop-words). Among the remaining tokens, we extract all the nouns and we reduce them to lemmas, then we compute the frequencies of the remaining tokens. This allow us to extract the scene-types (\textit{restaurant, beach} and \textit{airport}) from the captions, such as those shown in the examples above. The whole procedure is performed using spaCy. \footnote{\url{https://pypi.org/project/spacy/}}

\paragraph{Rank Biased Overlap}
RBO \cite{webber2010similarity} computes the similarity of two ranked lists, as follows:
\begin{equation}
    RBO(S, T, p) = (1-p)\sum p^{d-1}A_d
\end{equation}
\noindent
where $d$ is the depth of the ranking being examined, $A_d$ is the agreement between $S$ and $T$ given by the proportion of the size of the overlap up to $d$, and $p$ determines the contribution of the top $d$ ranks to the final RBO measure. We use the standard value of $p=1$.

\begin{table}[h!]
\centering
\begin{tabularx}{\columnwidth}{X|X|X}
    \hline
    \# Ablation & Train-Val & Test \\\hline \hline
    no ablation & 13498 & 1499 \\ \hline
    1 & 4269 & 469 \\ 
    2 & 2565 & 274 \\ 
    3 & 1554 & 170 \\ \hline
    \end{tabularx}
    \caption{\label{ablation_splits} Sample size of the Train-Val and Test split after ablation of the top 1,2 and 3 most informative objects in the most frequent scenes. The top row corresponds to the original dataset split sizes.}
\end{table}

\section{Ablation Details}
\label{sec:app-ablation}
As described in Section~\ref{sec:ablation} the ablation is performed by removing the most informative objects from images depicting the most frequent scene types. As a result, an image is included in the ablation study if (i) it belongs to the set of most frequent scenes; and (ii) it contains the objects we want to ablate. This means that the higher the number of objects ablated, the smaller the sample of images matching these constraints. As shown in Table~\ref{ablation_splits}, with 3 objects ablated in the test-set we obtain 170 valid images.

\begin{figure*}[]
\centering
    \includegraphics[scale=0.55]{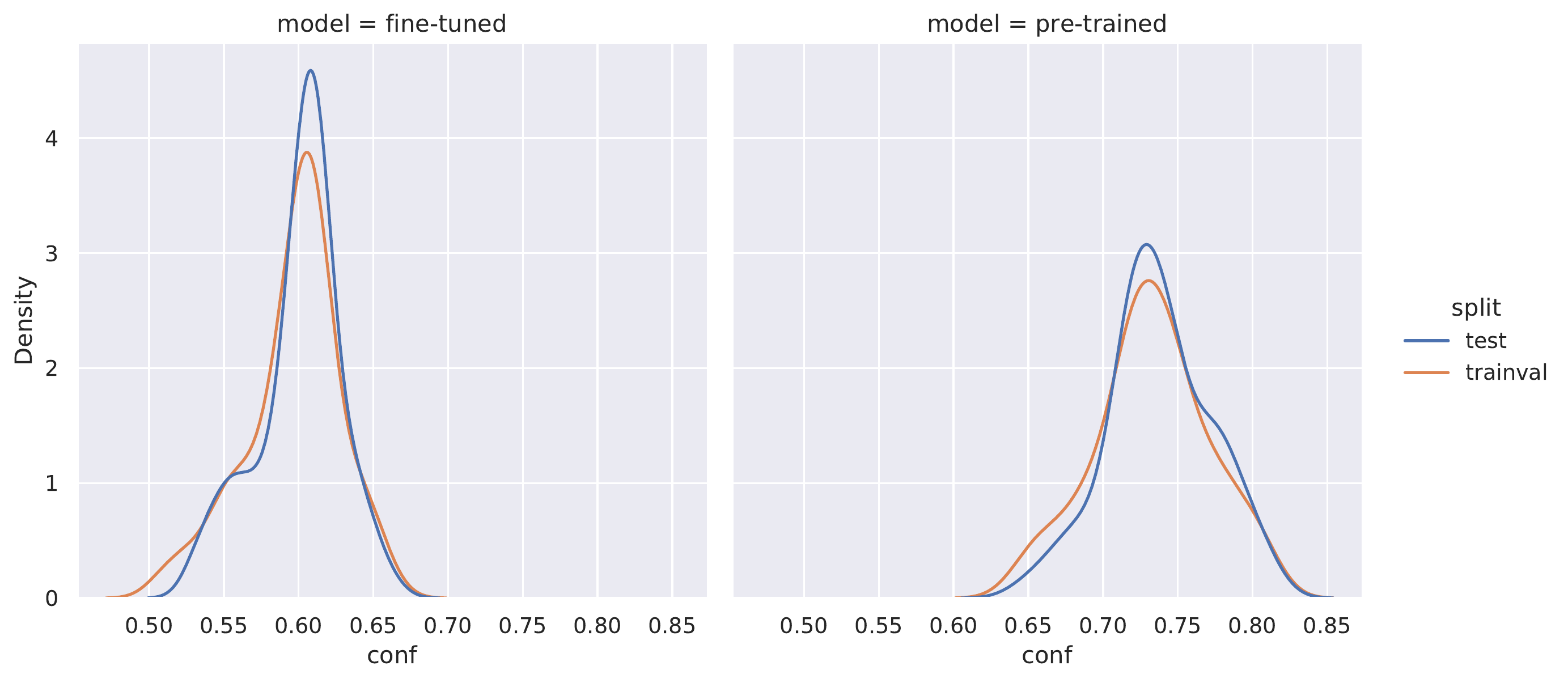}
    \caption{Kernel density estimate of the confidence scores distributions of unchanged captions after ablation for the test (blue) and train-val (orange) split. }
    \label{fig:conf_split}
\end{figure*}

We repeat the ablation experiment on both the test and the train-val split. The results obtained on the latter mirror those reported in Section~\ref{sec:ablation} with the test-split only. In Figure~\ref{fig:conf_split} we show the comparison of the distributions of the unchanged confidence scores after ablation for the test and train-val split. Moreover, there is no statistically significant difference between the distributions of confidence score shifts of the test set (shown in Figure~\ref{fig:violin}) and the train-val set ($z = 0.13$ with $p=0.89$ and $\alpha=0.05$).

\section{Probing details}
\label{sec:prob_details}

\paragraph{Model selection}
We test two probing models: a multi-layer perceptron and a random forest. We perform hyperparameter tuning of the neural probe by carrying out a random search followed by a probabilistic search. The tuned neural probe is a three-layer feed-forward network with \textit{hidden size} 16, optimized using LBFGS with adaptive learning rate and $\alpha=1$. Note that no parameter tuning is required for the random forest. As reported in Table~\ref{probe_f1_full}, the random forest performs better or on a par with the neural probe. Therefore we report the performance of the random forest in the main results in Section~\ref{sec:prob}.
\begin{table}[!h]
\small
\begin{tabularx}{\columnwidth}{X|X|XXX}
\hline
Probe  & Model  & micro-F1 & macro-F1 & weighted-F1 \\ \hline \hline
 \multirow{1}{*}{RB} &                        & 0.16    & 0.12      & 0.16 \\ \hline \hline
 \multirow{4}{*}{RF} &   PRE            & 0.94     & 0.67     & \bf{0.92}        \\ 
                     &   FT             & \bf{0.99}     & \bf{0.96}     & \bf{0.99}        \\ 
                     &   PRE (A) & 0.92     & 0.66     & 0.90        \\ 
                     &   FT (A)  & 0.98     & \bf{0.88}     & 0.97        \\ 
\hline
 \multirow{4}{*}{MLP} &  PRE            & 0.94     & 0.67     & 0.91        \\ 
                     &   FT             & 0.98     & 0.91     & 0.98        \\ 
                     &   PRE (A) & 0.92     & 0.66     & 0.90        \\ 
                     &   FT (A)  & 0.98     & 0.85     & 0.97        \\ 
\hline
\end{tabularx}

\caption{\label{probe_f1_full} F1-scores of scene classification task in the 50/50 split, for Random Baseline (RB), Random Forest (RF) and Multilayer perception (MLP) trained on encodings extracted from the pre-trained (PRE) and fine-tuned (FT) model without and with ablation (A). In bold the best result for each setting.}
\end{table}

\begin{figure}[t]
\centering
    \includegraphics[scale=0.4]{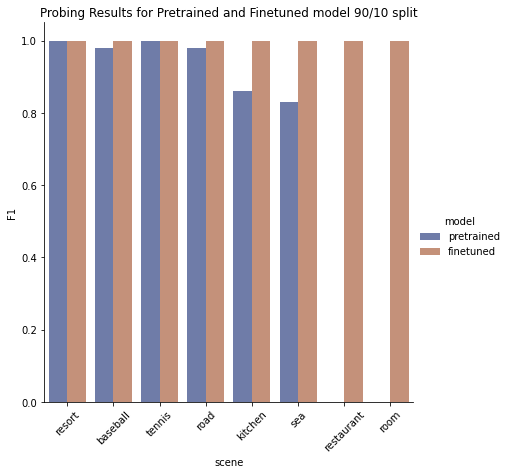}
    \caption{F1-scores of the scene classification task for the pre-trained (blue) and the fine-tuned model (orange) for the 90/10 split.}
    \label{fig:probe_ft_pre_10}
\end{figure}

\paragraph{Challenging the probe}
 The probing model performs at ceiling with the more typical 90/10 split, especially when trained on the fine-tuned features (Figure~\ref{fig:probe_ft_pre_10}). Therefore, we perform multiple experiments for different train/test splits namely, 90/10, 70/30 and 50/50. The 50/50 is the most challenging for the probe and it allows us to highlight the performance gap across different settings. Results from all the splits are shown in Table \ref{probe_f1_splits}. 
 
\begin{table*}[]

\centering
\begin{tabularx}{\linewidth}{X|X|XXX}
\hline
Split  & Model  & micro-F1 & macro-F1 & weighted-F1 \\ \hline \hline
 \multirow{4}{*}{90/10} &  PRE            & 0.96     & 0.71     & 0.94        \\ 
                     &   FT             & \bf{1.0}     & \bf{1.0}     & \bf{1.0}        \\ 
                     &   PRE (A) & 0.95     & 0.69     & 0.94        \\ 
                     &   FT (A)  & 0.99     & 0.99     & 0.99        \\ 
\hline
 \multirow{4}{*}{70/30} &  PRE            & 0.94     & 0.67     & 0.92        \\ 
                     &   FT             & \bf{0.99}     & \bf{0.97}     & \bf{0.99}        \\ 
                     &   PRE (A) & 0.93     & 0.66     & 0.91        \\ 
                     &   FT (A)  & 0.98     & 0.94     & 0.98        \\ 
\hline
 \multirow{4}{*}{50/50} &  PRE            & 0.94     & 0.67     & 0.92        \\ 
                     &   FT             & \bf{0.99}     & \bf{0.96}     & \bf{0.99}        \\ 
                     &   PRE (A) & 0.92     & 0.66     & 0.90        \\ 
                     &   FT (A)  & 0.98     & 0.88     & 0.97        \\ 
\hline
\end{tabularx}

\caption{\label{probe_f1_splits} F1-scores for scene classification task the random forest in different train/tes splits. The random forest is trained on encodings extracted from the pre-trained (PRE) and fine-tuned (FT) model without and with ablation (A).}
\end{table*}

\newpage
\section{HL-Scences examples}
\label{sec:app-hl}
\begin{table*}[t]
    \small
    \centering
    
    \begin{tabular}{b{.5\linewidth}b{.2\linewidth}b{.2\linewidth}}
        
        \bf Image  & \bf Object description (COCO) & \bf Scene description (HL-Scenes) \\
        \toprule
        \includegraphics[width=.75\linewidth]{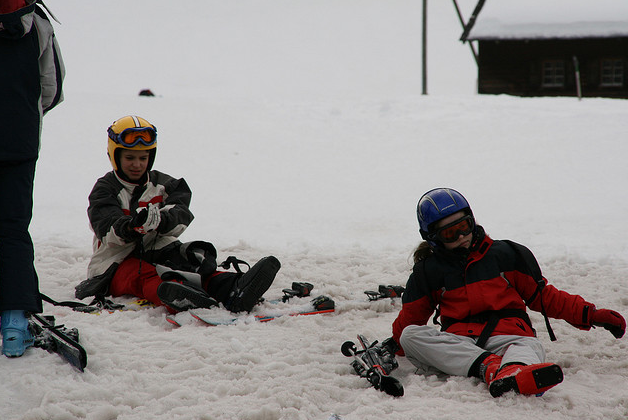}
        & a woman and a boy sitting in the snow outside of a cabin.  & the picture is shot in a ski resort \\
        \midrule
        \includegraphics[width=.75\linewidth]{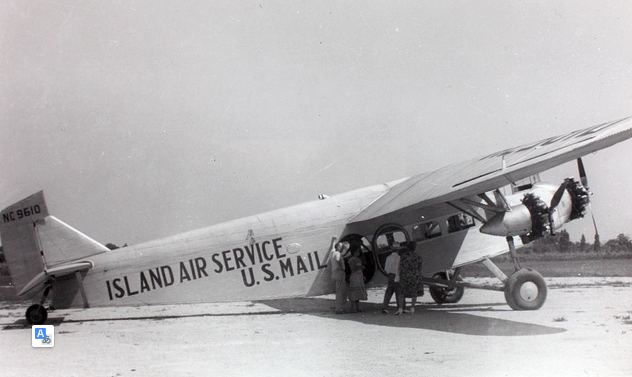}
        & a airplane with a group of people standing next to it.  & the picture is shot in an airport \\
        \midrule
        \includegraphics[width=.75\linewidth]{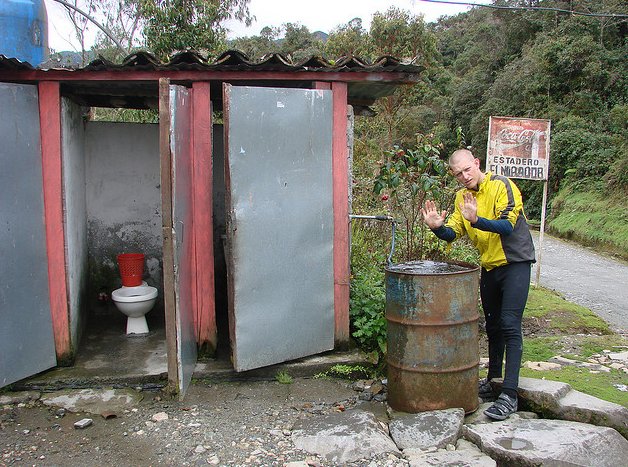}
        & a man holds his hands up as he stands over a trash can.  & the picture is taken in front of a roadside toilet \\
        \midrule
        \includegraphics[width=.75\linewidth]{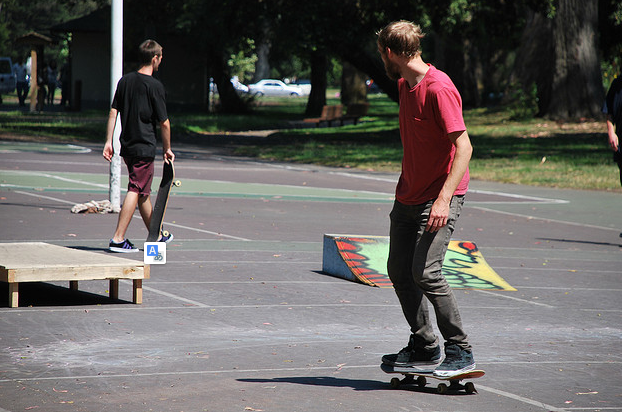}
        & a coupe of people that are skateboarding on a ramp  & it is at the park. \\
        \bottomrule
      
    \end{tabular}
    \caption{Randomly selected images from the HL-scenes dataset. For both COCO and HL-Scenes we show a randomly picked caption among the the available ones for the image.}
    \label{tab:hl-examples}
\end{table*}

\end{document}